\documentclass[12pt]{article}
\usepackage{times, epsfig, amsmath, amssymb, amsfonts}
\usepackage{natbib}
\usepackage{xcolor}

\begin{document}
\thispagestyle{empty}

\begin{center}{{\Large \bf Are Deep Neural Networks ``Robust"?}
\vspace{0.3cm}

{\large Peter Meer}\\
Distinguished Professor Emeritus \\
Electrical and Computer Engineering Department\\
Rutgers University, New Jersey, USA\\
meer@soe.rutgers.edu} 
\end{center}

\vspace{0.4cm}

Robust estimation discards outliers --- the unwanted data --- and finds the inliers, the good data. This simple definition of robustness is wide-ranging, perhaps leading to its misuse. Research using deep neural networks frequently uses the word ``robustness" in the title (a Google Scholar search for ``deep neural network" and ``robust*" returns 31,200 results in the field of computer vision).  This essay argues that the robustness of deep neural networks is limited, and does not meet the typical definition used in computer vision.

The traditional approach of robust estimation models the inlier structures with mathematical functions. The estimators do not need training, as the results are established based on predefined algorithms. The input data can be modified without any changes in the main program. The output of the estimator can be used directly as the input into the next procedure. 

Researchers have long tried to reduce or even eliminate those input points which do not fit a potential estimate. As \citet[p.278]{stigler10} recalled, John W. Tukey wrote in 1960 that the estimation of the scale of the normal distribution is less efficient if a few values are contaminated a small amount at a distance of three standard deviations. \cite{huber64} translated this observation into the first robust statistical paper where contributions of the more distant points were systematically reduced.

The input data in computer vision are discrete and therefore assuming a standard distribution for the inliers cannot be completely true. Boasting that an estimate is close to the global optimum is misguided, since inputs that were not considered would likely give values far from the optimum. 

Statisticians are aware that the experimental procedures are at least as important as the chosen model, \citet[p.277]{morgenthaler07}. In computer vision, many predefined functions for the inliers exist. Therefore, the main problem is to first eliminate the outliers. We begin with a very old example,  predating even least squares estimation. 

Around 1749, the German cartographer and astronomer Tobias Mayer (1723--1762) derived 27 equations to study the orbit of the moon from observations of a crater on the moon. He used the method of averages, summing up groups of nine equations into a new equation. ``The differences between the three sums are made as large as possible" and in this simple case, the three unknowns were more accurately found, \citet[pg.16--25]{stigler86}, \citet[p.44]{hald07}.

The Croatian polymath Roger Joseph Boscovich (1711--1787) published a method to measure the ellipticity of the earth's oblate shape in 1755. With measurements from five locations, he obtained solutions for two unknowns in all ten pairs. The average of the solutions was incorrect and he removed two ``so different from the others". The average of the eight remaining solutions was satisfactory, \citet[pg.39--50]{stigler86}, \citet[pg.45--46]{hald07}.

Both Mayer and Boscovich reduced the scope of the estimation to the number of unknown. But Mayer used sums, while Boscovich computed the minimum number of points needed to solve the problem. Boscovich also eliminated two pairs. These were the forerunners of elemental subsets and the removal of data above a threshold.

In computer vision, the first robust estimation paper was by \citet{fischler81}. The algorithm is called RANdom SAmple Consensus (RANSAC), and was similar to Boscovich's method.

The RANSAC estimators associate a linearized function derived from the nonlinear input with the inliers. Each term in the linearized function is a separate variable. The user gives two parameters before the estimation. Since the total number of elemental subsets in real problems is very large, the number $M$ establishes how many subsets will be used. The second parameter is $\sigma$, the scale of the inliers. The inliers' covariance matrix is the unit matrix multiplied with $\sigma$. 

\noindent
In RANSAC, the following procedure is repeated $M$ times, for $n$ data points:
\vspace{-0.2cm}
\begin{itemize}
\item Choose an elemental subset by random sampling without replacement. 
\vspace{-0.3cm}
\item Define a linear model candidate by the minimum number of points.
\vspace{-0.3cm}
\item Assume the candidate is valid for all $n$ points. Compute the distances between the points and the model. 
\vspace{-0.3cm}
\item Distances less than $\sigma$ give the inlier consensus set.
\vspace{-0.2cm}
\end{itemize}
The largest consensus set after $M$ trials, i.e., the smallest set of outliers, is the RANSAC estimate. Apply total least squares for the retained points and, if necessary, project the estimate back to the input.

Frequently, the user must give another parameter to end the algorithm, for example, when multiple inlier structures have to be detected. If the data contains too many outliers, a limiting value set-up by the input, RANSAC-type estimators often fails completely. 
Failure can also happen when images resized but the scale is not adjusted; from a sequence of images where the scale is significantly changed; from asymmetric outliers, etc. RANSAC-type estimators also do not work if the different inlier structures have very different noise processes. In the last 40 years, hundreds of paper have been written which tried to improve on the original RANSAC. But problems have always remained. 

\cite{yang20} recently published a new algorithm, the Multiple Input Structures with Robust Estimator (MISRE). The $M$ elemental subsets and linearized inlier model are still used but the user {\it no longer} specifies the inlier scale. There are several advantages relative to RANSAC-type estimators.\\
-- No difference exist between the treatment of the inlier structures or outliers.\\
-- Each iteration is independent from the other iterations.\\
-- Scale estimations have two constants, not parameters, given by the user.\\ 
-- The constants are the same for all estimation in computer vision.\\
-- The constants only weakly influence the scale estimates. \\
-- Stronger inlier structures are still recovered even with excessive outliers.

\noindent
MISRE is a more universal approach to extract robust mathematical functions and performs as well as the RANSAC-type estimators if those use the correct parameters. But benchmark performance comparisons between those estimators and MISRE is a flawed experiment. MISRE does not need different parameters, but RANSAC-type estimators do, and also have to be adjusted for specific cases.

In general, robust estimators do not exactly recover every point belonging to an inlier structure. Such a solution is not worthwhile since the RANSAC-type estimators have data dependent thresholds, while MISRE does not try to be that accurate. For a more precise solution with MISRE, post-processing with specific thresholds is needed.

Machine learning in general and computer vision in particular also have another definition for robustness based on deep neural networks with many layers and millions of parameters. Consecutive layers integrate small parts of an image. For the training sequence, the user first must decide which are the categories needing robust responses at the output. For example, if the subject is animals, then each category (dogs, cats, etc.) is a separate response. An image can have several examples in a category, each giving a response of one if detected. If the category does not exist or is not found, the response is zero. The user's interaction is limited to providing the input and accepting the output; the parameters for estimation are a black box. 

The network learns the parameters through a simple iterative procedure. Errors are summed and backpropagation updates the parameters to compute a new fit. There are many iterations in the training, which can take several days. Users stop the training based on a minimum validation threshold for errors. 

The outputs are usually complex figures for which no simple mathematical description exist. An output may also be outliers surrounding an inlier structure, since the inlier vs. outlier separation is not maintained.

Neural networks are brittle. A small adversarial perturbation of the input, perhaps undetectable by the user, can generate an incorrect output. Vint \citet{cerf20} explains that this issue arises because the way the networks are built. 

The test sequence is much smaller that the training sequence, but must to be similar. Based on responses from the training sequence network, the robust outputs are constructed. When applied correctly for a well-defined problem, deep neural networks can have better results than any previous approach. \cite{su18} showed that the correct estimate is returned first in more that 80\% of cases.

Comparison of RANSAC-type estimators with deep neural networks is not entirely fair. 
For example, the application of neural networks to classical computer vision tasks by \cite{brachmann19} shows that the outputs improve only a few percent, if at all. 

But deep neural networks have a major flaw.  If an input that is different than the training sequence is presented, the output is no longer correct. The whole network must be retrained if a new category is introduced.

In essence, therefore, the outputs are more artifacts of a long training session than robust estimates obtained directly from the input. The estimates cannot be called robust using the traditional definition laid out above, in which outliers in the data are separated from inliers.

The human vision system is amazingly robust. \cite{kruger13} describes the state of knowledge on the human vision system at the beginning of 2010s. While the fastest neurons have the latency of at least 20ms, one million times slower an a switch in a computer, 
the vision system achieves high efficiency in moving from a task to another one.
Deep neural networks can achieve performance better than human in some tasks, but this does not mean that processing is robust in the traditional sense.

The answer to the question in the title of the paper is ``no."

\bibliographystyle{plainnat}  
\bibliography{whatwehave}

\begin{thebibliography}{11}
\providecommand{\natexlab}[1]{#1}
\providecommand{\url}[1]{\texttt{#1}}
\expandafter\ifx\csname urlstyle\endcsname\relax
  \providecommand{\doi}[1]{doi: #1}\else
  \providecommand{\doi}{doi: \begingroup \urlstyle{rm}\Url}\fi

\bibitem[Brachmann and Rother(2019)]{brachmann19}
Eric Brachmann and Carsten Rother.
\newblock Neural-guided {RANSAC: L}earning where to sample model hypotheses.
\newblock In \emph{ICCV'19}, pages 4322--4331, 2019.

\bibitem[Cerf(2020)]{cerf20}
Vint Cerf.
\newblock Will computers ever think like human beings?
\newblock \url{https://www.youtube.com/watch?v=J63mKverb8w} On-line from, May
  21, 2020.
\newblock The Royal Institution, London.

\bibitem[Fischler and Bolles(1981)]{fischler81}
Martin~A. Fischler and Robert~C. Bolles.
\newblock Random sample consensus: A paradigm for model fitting with
  applications to image analysis and automated cartography.
\newblock \emph{Comm. Assoc. Comp. Mach.}, 24:\penalty0 381--395, 1981.

\bibitem[Hald(2007)]{hald07}
Anders Hald.
\newblock \emph{A history of parametric statistical inference from Bernoulli to
  Fisher, 1713 to 1935.}
\newblock Springer, 2007.

\bibitem[Huber(1964)]{huber64}
Peter~J. Huber.
\newblock Robust estimation of a location parameter.
\newblock \emph{Annals of Mathematical Statistics}, 35:\penalty0 73--101, 1964.

\bibitem[Kr\"{u}ger et~al.(2013)Kr\"{u}ger, Janssen, Kalkan, Lappe, Leonardis,
  Piater, Rodrı\'{i}guez-S\'{a}nchez, and Wiskott]{kruger13}
Norbert Kr\"{u}ger, Peter Janssen, Sinan Kalkan, Markus Lappe, Ale\u{s}
  Leonardis, Justus Piater, Antonio~J. Rodrı\'{i}guez-S\'{a}nchez, and Laurenz
  Wiskott.
\newblock Deep hierarchies in the primate visual cortex: {W}hat can we learn
  for computer vision?
\newblock \emph{IEEE Trans. Pattern Anal. Mach. Intel.}, 35:\penalty0
  1847--1871, 2013.

\bibitem[Morgenthaler(2007)]{morgenthaler07}
Stephan Morgenthaler.
\newblock A survey of robust statistics. {D}iscussion.
\newblock \emph{Stat. Meth. \& Appl.}, 15:\penalty0 271--293, 2007.

\bibitem[Stigler(1986)]{stigler86}
Stephen~M. Stigler.
\newblock \emph{The History of Statistics. {T}he Measurement of Uncertainty
  before 1900.}
\newblock Harvard University Press, 1986.

\bibitem[Stigler(2010)]{stigler10}
Stephen~M. Stigler.
\newblock The changing history of robustness.
\newblock \emph{The American Statistician}, 64:\penalty0 277--281, 2010.

\bibitem[Su et~al.(2018)Su, Zhang, Chen, Yi, Chen, and Gao]{su18}
Dong Su, Huan Zhang, Hongge Chen, Jinfeng Yi, Pin-Yu Chen, and Yupeng Gao.
\newblock Is robustness the cost of accuracy? {A} comprehensive study on the
  robustness of 18 deep image classification models.
\newblock In \emph{ECCV'18}, volume LNCS:1216, pages 644--661. Springer, 2018.

\bibitem[Yang et~al.(2020)Yang, Meer, and Meer]{yang20}
Xiang Yang, Peter Meer, and Jonathan Meer.
\newblock A new approach to robust estimation of parametric structures.
\newblock \emph{IEEE Trans. Pattern Anal. Mach. Intel.}, 42:\penalty0 Early
  Access, 2020.

\end{thebibliography}

\end{document}